\documentclass[letterpaper, 10 pt, journal, twoside]{ieeetran}  

\usepackage[nocompress]{cite}
\usepackage[pdftex]{graphicx}
\usepackage{color}
\usepackage{ragged2e}
\usepackage{multirow}
\usepackage{multicol}
\usepackage{amsmath,amssymb}
\usepackage[caption=false]{subfig}
\usepackage{threeparttable}
\usepackage{diagbox}
\usepackage{comment}
\usepackage{url}

\usepackage[colorlinks,bookmarksnumbered,bookmarksopen,linkcolor=black,citecolor=black,urlcolor=black]{hyperref}

\graphicspath{{./Imgs/}}
\DeclareGraphicsExtensions{.pdf,.jpg,.png}

\newcommand{\figref}[1]{Fig.~\ref{#1}}
\newcommand{\tabref}[1]{Tab.~\ref{#1}}
\newcommand{\secref}[1]{Sec.~\ref{#1}}

\newcommand{\equref}[1]{~(\ref{#1})}

\DeclareMathOperator{\arctantwo}{arctan2}

\def\ie{\textit{i.e.,~}}

\def\etal{\textit{et al.~}}
\def\sArt{{state-of-the-art~}}
\newcommand{\todo}[1]{#1}

\IEEEoverridecommandlockouts                              

\title{\LARGE \bf
Multi-scale Interaction for Real-time LiDAR Data Segmentation on an Embedded Platform
}

\author{Shijie Li, 
        Xieyuanli Chen, 
        Yun Liu, 
        Dengxin Dai, 
        Cyrill Stachniss, 
        Juergen Gall,~\IEEEmembership{Member,~IEEE}
\thanks{Manuscript received: June 9, 2021; Revised October 6, 2021; Accepted November 18, 2021. This paper was recommended for publication by Editor Markus Vincze upon evaluation of the Associate Editor and Reviewers' comments.}
\thanks{S. Li, X. Chen, C. Stachniss, and J. Gall are with the University of Bonn, Germany. D. Dai is with MPI for Informatics, Germany. Y. Liu is with ETH Zurich, Switzerland. This work was supported by the Deutsche Forschungsgemeinschaft (DFG, German Research Foundation) under Germany's Excellence Strategy - EXC 2070 - 390732324 and GA1927/5-2 (FOR 2535 Anticipating Human Behavior).}%
}

\begin{document}

\maketitle

\begin{abstract}
Real-time semantic segmentation of LiDAR data is crucial for autonomously driving vehicles and robots, which are usually equipped with an embedded platform and have limited computational resources. Approaches that operate directly on the point cloud use complex spatial aggregation operations, which are very expensive and difficult to deploy on embedded platforms. As an alternative, projection-based methods are more efficient and can run on embedded hardware. However, current projection-based methods either have a low accuracy or require millions of parameters. In this paper, we therefore propose a projection-based method, called Multi-scale Interaction Network (MINet), which is very efficient and accurate. The network uses multiple paths with different scales and balances the computational resources between the scales. Additional dense interactions between the scales avoid redundant computations and make the network highly efficient. The proposed network outperforms point-based, image-based, and projection-based methods in terms of accuracy, number of parameters, and runtime. \todo{Moreover, the network processes more than 24 scans, captured by a high-resolution LiDAR sensor with 64 beams, per second on an embedded platform, which is higher than the framerate of the sensor}. The network is therefore suitable for robotics applications.


\end{abstract}

\section{Introduction}\label{sec:introduction}

Environment perception and understanding are key to realize self-driving vehicles and robots.
For the full-view perception of the environment, autonomously driving vehicles are usually equipped with multi-sensor systems, among which light detection and ranging (LiDAR) sensors play a key role due to their precise distance measurements. The large point clouds that are generated by the LiDAR sensors, however, need to be interpreted in order to understand the environment.

Although convolution neural networks (CNNs) perform well for semantic image segmentation \cite{chen2017rethinking,zhao2017pyramid,howard2019searching,ma2018shufflenet}, they cannot be applied directly to 3D point clouds.
This is because standard convolutions require a regular grid structure, whereas a raw point cloud is an unordered structure.
To address this problem, some methods
\cite{qi2017pointnet,qi2017pointnet++,wu2019pointconv} directly process point clouds using some spatial aggregation operations like grouping and gathering.
Although these methods work well in indoor scenarios, it is difficult to apply them to large outdoor scenarios since the computational cost of the aggregation operation increases with the number of points. Another issue is that these methods are inefficient on embedded platforms since they use operations that cannot be efficiently mapped on embedded hardware like Jetson AGX using TensorRT. However, runtime efficiency is of vital importance for real-world applications, especially for autonomously driving vehicles and robots.

Wu \etal\cite{wu2018squeezeseg,wu2019squeezesegv2} thus proposed to represent point clouds produced by a LiDAR sensor as an ordered projection map, such that CNNs can then be applied. However, projected LiDAR data and RGB images are different modalities and applying directly 2D image-based methods does not yield a high segmentation accuracy.   
For this reason, some specific CNNs have been designed for LiDAR-based depth images, named as projection-based methods.
Recent projection-based methods like \cite{milioto2019rangenet++}, however, are very large with more than 50M parameters, making them not suitable for embedded platforms.


\begin{figure*}[t]
    \centering
    \includegraphics[width=0.9\linewidth]{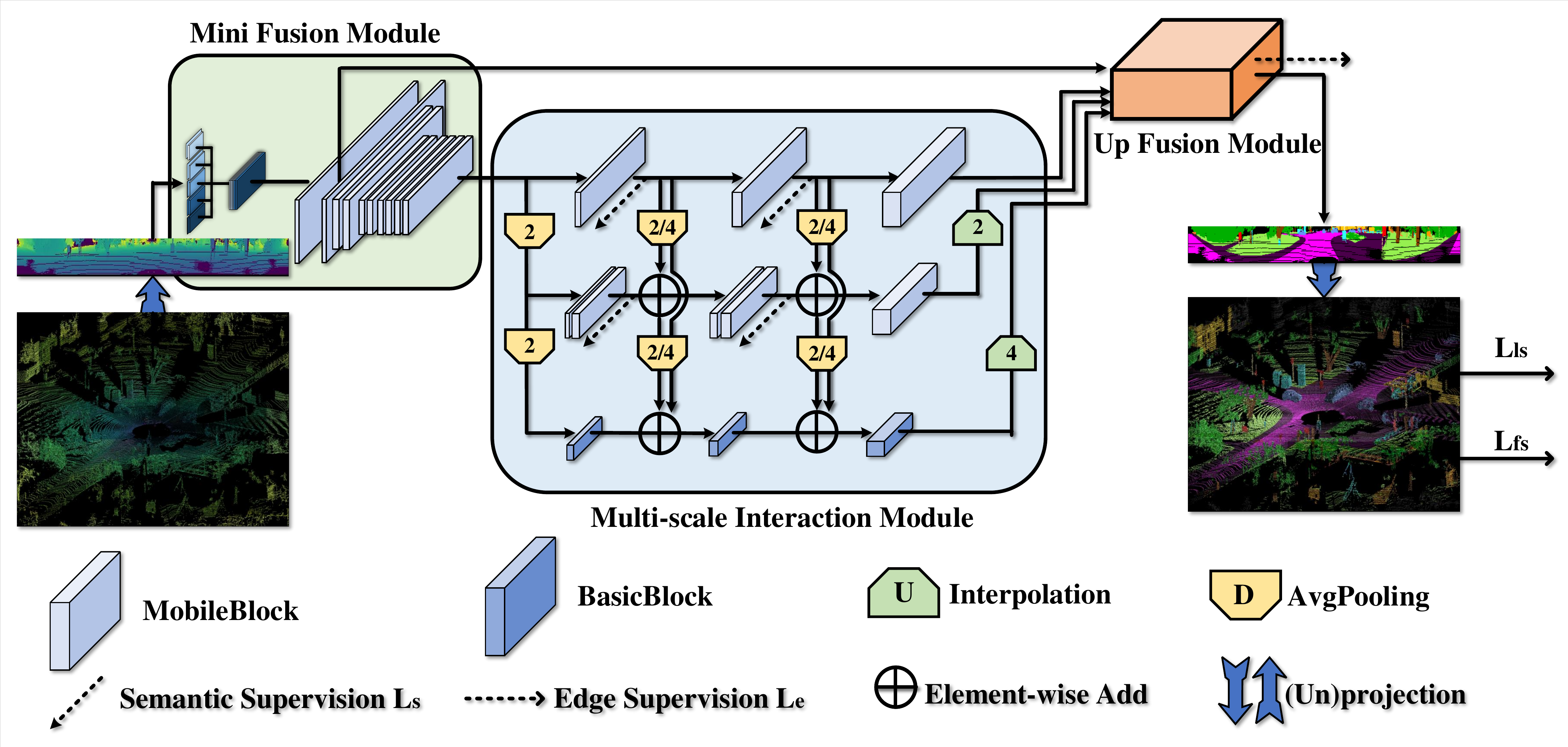}
    \caption{\todo{Illustration of the MINet architecture with three paths in the Multi-scale Interaction Module.} The numbers 2 and 4 for interpolation (U) and average pooling (D) indicate the upsampling and downsampling factor. The dashed arrows indicate the supervision type. \todo{The detailed description of the architecture is given in \tabref{tab:instantiation} where the different blocks are illustrated in \figref{fig:blocks} and the Up Fusion Module is illustrated in \figref{fig:ufm}.}}
    \label{fig:arch}
    \vspace{-4mm}
\end{figure*}

In this work, we therefore propose a lightweight projection-based model for semantic segmentation of LiDAR data that runs in real-time on an embedded platform. To this end, we revisit common multi-scale approaches like U-Net \cite{ronneberger2015u} that have one path for each scale\todo{, \ie each scale is processed independently and then fused at the end of the network.} 
These networks, however, use the same operations for each path, which makes them either too expensive for embedded platforms or the accuracy is very low depending how complex the used operations are. In order to achieve a good balance between effectiveness and efficiency, we therefore adapt the computational operations for each path.
While the top path extracts low-level clues, which can be easily detected with shallow layers operating on high-resolution feature maps, the bottom path extracts high-level semantic information, which requires more complex operations but on low-resolution feature maps. In order to avoid redundant computations across the paths, we furthermore propose a dense top-to-bottom interaction strategy where feature maps from a path are passed to all lower paths. We term the network, which is shown in \figref{fig:arch}, Multi-scale Interaction Network (MINet). 

In addition, we show that the accuracy can be increased if additional supervision is added. While this is consistent with \cite{zhao2017pyramid,zhang2018context,liu2019richer,liu2018semantic}, we demonstrate that it is important to use the right type of supervision for the right part of the network. In fact, we use semantic supervision only for the two top paths but not for the bottom path and edge supervision for the fusion of the multiple paths. The latter is important to obtain accurate segment boundaries after upsampling the paths with lower resolution.
Finally, we process the multi-modal data consisting of 3D coordinates, remission, and depth information first independently and then fuse them in the feature space. This is in contrast to previous works for LiDAR data that just concatenate the modalities and therefore ignore that the characteristics of each modality are different.

In summary, our contributions include:
\begin{itemize}
    \item We propose a multi-scale approach where the computational operations are balanced across the different scales and a top-to-bottom interaction strategy avoids redundant computations. 
    \item We exploit different types of additional supervision to improve the accuracy without increasing the inference time.
    \item Different from previous methods, we process each modality independently and fuse them in the feature space, which improves the overall segmentation performance.
    \item By incorporating the above design decisions, we propose a lightweight projection-based model for semantic segmentation of LiDAR data that runs in real-time on an embedded platform.
\end{itemize}
The experimental results demonstrate that our method reduces the number of parameters by about 98\% and is about 4$\times$ faster than the \sArt projection-based method \cite{milioto2019rangenet++}, while achieving a higher accuracy.
We also evaluate our model on an embedded platform and demonstrate that our method can be deployed for autonomous driving.

\section{Related Work}

\subsection{Point-based Semantic Segmentation}
Although CNNs \cite{chen2017rethinking,zhao2017pyramid,howard2019searching,ma2018shufflenet} are successful for 2D image-based semantic segmentation, they cannot handle unstructured data like point clouds.
To address this problem, tangent convolutions~\cite{tatarchenko2018tangent} 
project local points to a tangent plane and vanilla convolutions are then applied to it.
PointNet~\cite{qi2017pointnet} is the first method that directly 
processes the point cloud.
It applies a convolution operation for each point and uses a permutation invariant operation to aggregate information.
However, PointNet does not take local information into consideration, which is realized by PointNet++~\cite{qi2017pointnet++}.
SPGraph \cite{landrieu2018large} tackles semantic segmentation of large-scale point clouds by defining a super point graph (SPG).
Because point-based methods are inefficient for large point clouds, RandLA \cite{hu2019randla} addresses this problem by adopting random sampling and designing a better grouping strategy to maintain a better performance.
P$^2$Net \cite{li2020projected} applies point-based methods on projected LiDAR data.
However, these methods are too expensive for many applications, especially for embedded platforms.

\subsection{Projection-based Semantic Segmentation}\label{sec:relwork_proj}
Projection-based segmentation methods project LiDAR point clouds onto 2D multi-modal images and use 2D CNNs for semantic segmentation. SqueezeSeg~\cite{wu2018squeezeseg} and 
SqueezeSegV2~\cite{wu2019squeezesegv2} 
use a lightweight network called SqueezeNet~\cite{iandola2016squeezenet}
for semantic segmentation and a CRF for post-processing.
Based on SqueezeSeg,  RangeNet++~\cite{milioto2019rangenet++} adopts Darknet~\cite{redmon2018yolov3} and replaces the CRF with a $k$-NN for post-processing.
It has also been successfully used to improve LiDAR-based odometry~\cite{chen2019suma++} and loop closure detection~\cite{chen2020overlapnet}.
Current projection-based methods, however, do not achieve the same segmentation accuracy as point-based methods and the best performing approaches use very large networks. In this paper, we propose a novel lightweight model that can run in real-time on an embedded platform while achieving \sArt performance.

\begin{table}[!t]
    \centering
    \renewcommand{\tabcolsep}{3.0mm}
    \caption{Instantiation of the proposed MINet.}
    \label{tab:instantiation}
    \resizebox{\linewidth}{!}{%
    \begin{threeparttable}%
    \begin{tabular}{c|ccccccc}
        \hline
        Module & Operation & $k$ & $c$ & $s$ & $t$ & Output size \\
        \hline
        \multirow{12}*{MFM} & Conv2d & 3 & 4 & 1 & 5 & 64$\times$2048 \\
         & MobileBlock & 3 & 20 & 1 & 1 & 64$\times$2048 \\
         & MobileBlock & 3 & 24 & 2 & 1 & 32$\times$512 \\
         & MobileBlock & 3 & 24 & 1 & 1 & 32$\times$512 \\
         & MobileBlock & 5 & 40 & 2 & 1 & 16$\times$256 \\
         & MobileBlock & 5 & 40 & 1 & 1 & 16$\times$256 \\
         & MobileBlock & 5 & 40 & 1 & 1 & 16$\times$256 \\
         & MobileBlock & 3 & 80 & 1 & 1 & 16$\times$256 \\
         & MobileBlock & 3 & 80 & 1 & 1 & 16$\times$256 \\
         & MobileBlock & 3 & 80 & 1 & 1 & 16$\times$256 \\
         & MobileBlock & 3 & 80 & 1 & 1 & 16$\times$256 \\
         & Conv2d & 1 & 32 & 0 & 1 & 16$\times$256 \\ \hline
        \multirow{11}*{MIM} & MobileBlock & 3 & 64 & 1 & 1 & 16$\times$256 \\
         & MobileBlock & 3 & 128 & 1 & 1 & 16$\times$256 \\
         & MobileBlock & 3 & 128 & 1 & 1 & 16$\times$256 \\ \cline{2-7}
         & MobileBlock & 3 & 32 & 1 & 1 & 8$\times$128 \\ 
         & MobileBlock & 3 & 64 & 1 & 1 & 8$\times$128 \\
         & MobileBlock & 3 & 64 & 1 & 1 & 8$\times$128 \\
         & MobileBlock & 3 & 128 & 1 & 1 & 8$\times$128 \\
         & MobileBlock & 3 & 128 & 1 & 1 & 8$\times$128 \\
        \cline{2-7}
         & BasicBlock & 3 & 64 & 1 & 1 & 4$\times$64 \\
         & BasicBlock & 3 & 128 & 1 & 1 & 4$\times$64 \\
         & BasicBlock & 3 & 128 & 1 & 1 & 4$\times$64 \\ \hline
        \multirow{3}*{UFM} & Conv2d & 3 & 32 & 1 & 1 & 16$\times$512 \\
         & Conv2d & 3 & 32 & 1 & 1 & 64$\times$2048 \\
         & Conv2d & 1 & 32 & 1 & 1 & 64$\times$2048 \\ \hline
    \end{tabular}
    \begin{tablenotes}
    \item[*] Each module contains several components: Conv2d, MobileBlock, and BasicBlock. 
    Conv2d denotes a convolutional layer followed by one batch normalization layer and ReLU activation.
    MobileBlock and BasicBlock are illustrated in \figref{fig:blocks}. 
    Each operation has a kernel size $k$, stride $s$, and $c$ output channels, repeated $t$ times.
    The three sections of MIM denote the three paths.
    \end{tablenotes}
    \end{threeparttable}}
    \vspace{-8mm}
\end{table}

\section{Multi-scale Interaction Network}
The proposed Multi-scale Interaction Network (MINet) operates on projection maps generated from LiDAR point clouds. \todo{To associate a LiDAR point $\mathbf{a}=(x, y, z)$ to a pixel $(u, v)$ in the projection map of size $(h, w)$, we compute yaw \eqref{eq:yaw} and pitch \eqref{eq:pitch} and map it to pixel coordinates by translation and
scaling \cite{milioto2019rangenet++}:
}
\begin{equation}\label{eq:yaw}
    u = \frac{1}{2}[1 - \arctantwo(y, x)\pi^{-1}]w,
\end{equation}
\begin{equation}\label{eq:pitch}
    v = [1 - (\arcsin(zd^{-1}) + o_{up})o^{-1}]h.
\end{equation}
The vertical field-of-view of the LiDAR sensor is $o = o_{up} + o_{down}$, where $o_{up}$ and $o_{down}$ represent the above and below horizon of the field-of-view, respectively.
$d = ||\mathbf{a}||$ denotes the depth of a point.
\todo{After this transformation, we obtain a projection map of size $(h, w, 5)$ where the 5 channels correspond to the coordinates $(x, y, z)$, the depth, and the remission of the corresponding 3D point.} The remission value indicates the proportion of the light that is diffusely reflected. It provides therefore information of the surface, which is helpful for distinguishing different classes. \todo{While depth can be computed from the coordinates, the network operations do not compute the depth explicitly. Adding depth in addition to the coordinates thus improves the accuracy as we show in our experiments.}
Each channel is normalized by the mean and standard deviation computed over the training and validation set.

\begin{figure}[!t]
    \centering
    \includegraphics[width=0.8\linewidth]{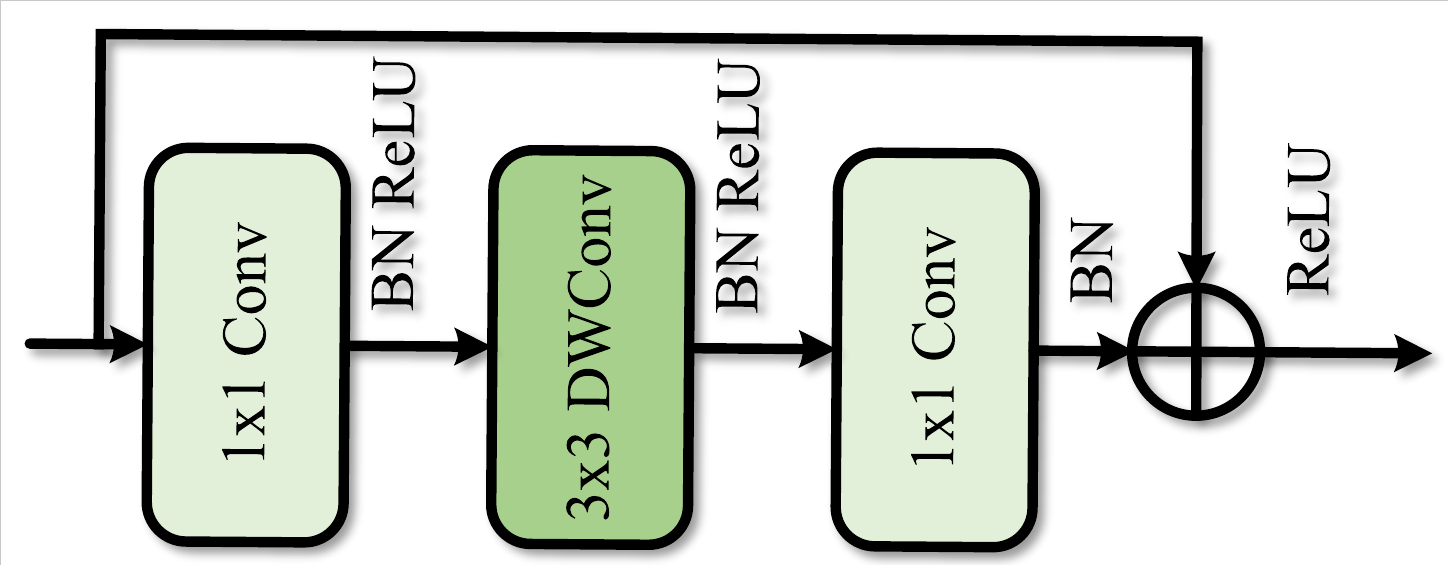}
    \\ \vspace{2mm}
    \includegraphics[width=0.8\linewidth]{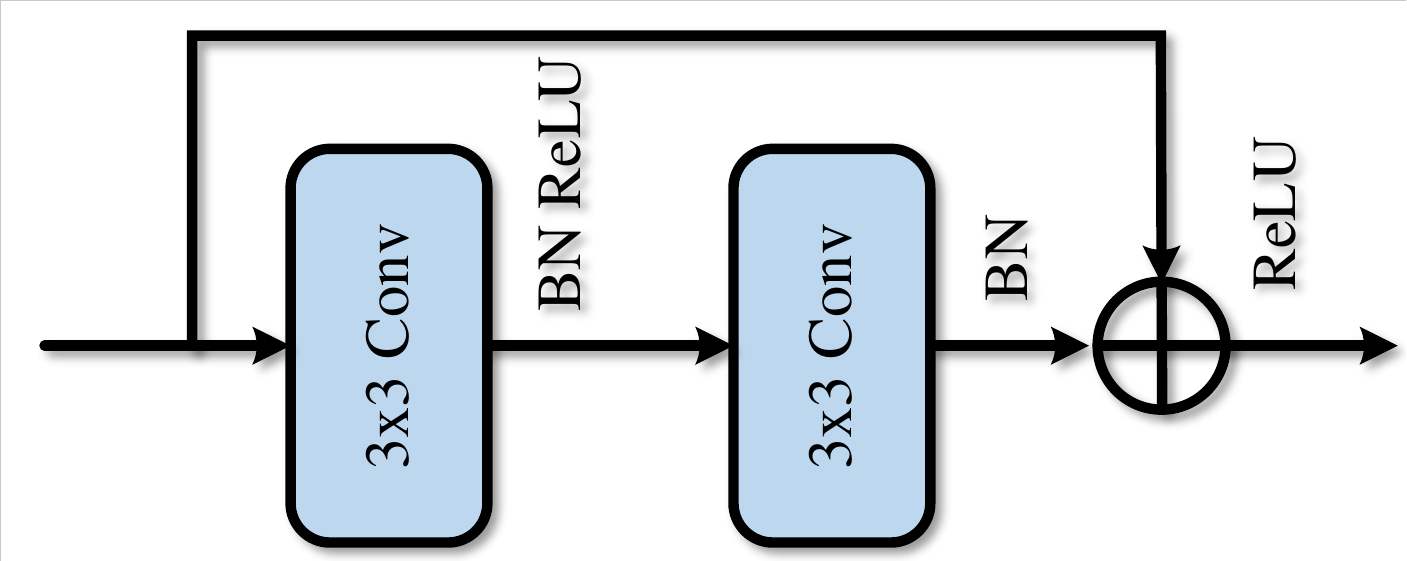}
    \caption{Illustration of the MobileBlock (top) and the BasicBlock (bottom). DWConv means depth-wise convolution.}
    \vspace{-6mm}
    \label{fig:blocks}
\end{figure}

The architecture of MINet is shown in \figref{fig:arch}. The projection map is first processed by the Mini Fusion Module (MFM) (\secref{sec:mfm}) to fuse the multi-modal information in the feature space. 
In the Multi-scale Interaction Module (MIM) (\secref{sec:mim}), the data is processed at three different scales where the resolution is reduced by factor two for each path. 
As it is shown in \tabref{tab:instantiation}, the computation differs for each path where we use two basic components, namely MobileBlock and BasicBlock. 
The MobileBlock \cite{howard2017mobilenets, sandler2018mobilenetv2} utilizes depthwise convolutions and has thus fewer parameters, but its learning capacity is also limited. 
The BasicBlock~\cite{he2016deep} is stronger, but also more expensive. 
MobileBlock and BasicBlock are shown in \figref{fig:blocks}. 
We therefore balance the computational resources across the three paths as it is shown in \tabref{tab:instantiation}. 
While we use the expensive BasicBlock for the bottom path with lowest resolution, we decrease the computational cost as the resolution increases using five MobileBlocks for the middle path and three for the top path. 
The connections from each path to lower paths avoid redundant computations at lower paths and make the network more efficient. 
Finally, the 2D predictions for the original resolution are produced by the Up Fusion Module (UFM) (\secref{sec:ufm}), which are then mapped back to the 3D space. 
In the remainder of this section, we describe each module of MINet.

\subsection{Mini Fusion Module (MFM)} \label{sec:mfm}
Different from an RGB image, the projection map contains channels of different modalities. 
Previous projection-based segmentation methods \cite{wu2018squeezeseg,wu2019squeezesegv2,milioto2019rangenet++} treat such different modalities equally, but we show that processing each channel independently using MFM is more efficient.
Specifically, each channel of the multi-modal image is mapped to an independent feature space using five convolution blocks, including normalization and activation. 
This corresponds to the first row of \tabref{tab:instantiation}. 
This step can be considered as a feature calibration step for each modality before fusing them. 
It needs to be noted that we also treat the $x$, $y$, and $z$ coordinates separately. 
After the first five convolutional layers, these features are concatenated and fed into several MobileBlocks for fusing them. 
Since a small resolution leads to less computation, the information of the feature maps are gradually aggregated by average pooling.

\subsection{Multi-scale Interaction Module (MIM)} \label{sec:mim}
After the fusion module, the data is processed by three paths where each path corresponds to a different scale as shown in \figref{fig:arch}. From top to bottom, the resolution of the feature maps is decreased by factor two using average pooling and the receptive field is accordingly increased. For the top path, we use the highest resolution. Since processing high resolution feature maps is very expensive, we use only three MobileBlocks as shown in \tabref{tab:instantiation}. The bottom path, which has the largest receptive field and lowest resolution, can offer more abstract semantic clues if we use more expensive operations. Hence, it uses three BasicBlocks. The middle path is a compromise between the top and bottom path and consists of five MobileBlocks. In our experiments, we show that increasing the computational operations as the resolution decreases leads to a higher efficiency compared to using the same blocks for all paths. While the number of parameters doubles compared to the proposed architecture if we use the BasicBlocks for all paths, the accuracy drops if only MobileBlocks are used.

A second important design choice is to allow interactions among the paths. 
Since the computational complexity of the paths increases for lower paths, we use a dense top-to-bottom fusion design for efficient multi-scale feature interaction. Especially, feature maps of the first and second path will be resized by average pooling and passed to all lower paths.
To avoid a mismatch of the number of channels, the number of channels is increased gradually for each path and kept the same at each interaction position. Hence, no other operations are used to adjust the number of channels as shown in \tabref{tab:instantiation}. Due to the interaction, the lower paths benefit from the features computed from higher paths. The lower paths can therefore focus on information that has not been extracted by higher paths due to limited computational resources.        

\begin{figure}[!t]
    \centering
    \includegraphics[width=.8\linewidth]{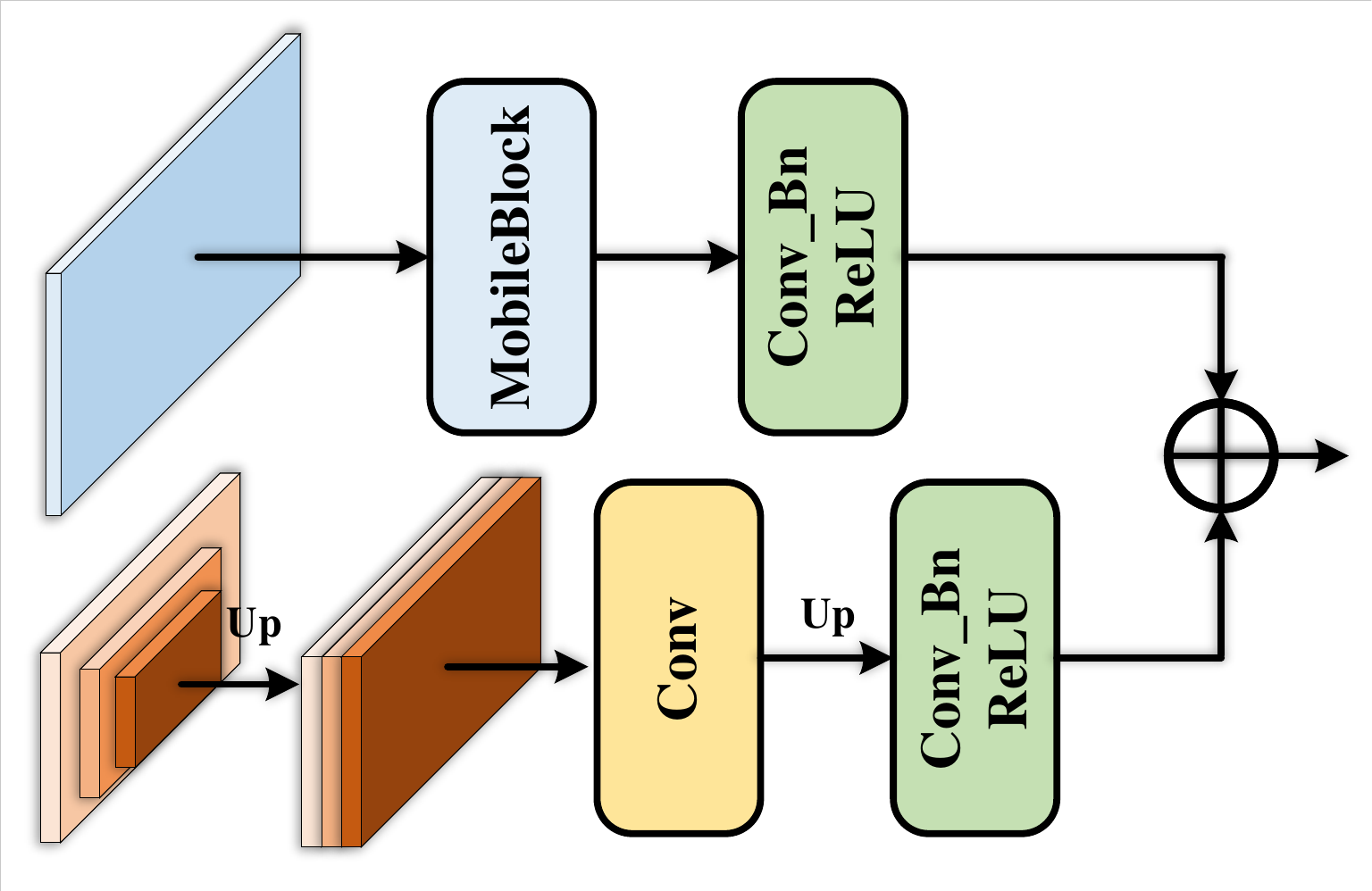}
    \caption{Illustration of the Up Fusion Module (UFM).}
    \label{fig:ufm}
    \vspace{-4mm}
\end{figure}

\begin{table}[!t]
    \centering
    \renewcommand{\tabcolsep}{3.5mm}
    \caption{Impact of the three modules.}
    \label{tab:ablation}
    \resizebox{\linewidth}{!}{%
    \begin{tabular}{c|ccc|c} \hline
        No. & MFM & Interaction & UFM & mIoU \\ \hline
        1  & & \checkmark & \checkmark & 50.9 \\
        2 & \checkmark & & \checkmark &  50.7 \\
        3  & \checkmark & \checkmark & & 50.6 \\
        4  & \checkmark & \checkmark & \checkmark & 51.8 \\
        5  & w/o depth & \checkmark & \checkmark & 49.6 \\
         \hline
    \end{tabular}}
    \vspace{-3mm}
\end{table}

\subsection{Up Fusion Module (UFM)} \label{sec:ufm}
To obtain the semantic labels of each pixel in the projection map, UFM shown in \figref{fig:ufm} combines the features from different scales and upsamples them to the input resolution. In addition, features after the first MobileBlock of the Mini Fusion Module are used to recover detailed spatial information as shown in \figref{fig:arch}. The lower part of \figref{fig:ufm} shows the feature maps of the three different paths that are first resized to the same size, concatenated together, and fused by a 1$\times$1 convolution. The fused feature maps are then upsampled to the original resolution and processed by a convolution block including a 3$\times$3 convolution, batch normalization, and ReLU activation. The upper part shows the feature maps from the Mini Fusion Module that have already the original resolution. They are processed by a MobileBlock and a convolution block. Finally, the processed features from both modules are added together. Although the spatial information of the original feature maps already helps to sharpen segment boundaries, which can be fuzzy due to the upsampling, adding additional supervision for the segment boundaries emphasizes this effect as we will explain in the next section.

\begin{table}[!t]
    \centering
    \renewcommand{\tabcolsep}{3.2mm}
    \caption{Impact of additional supervision.}
    \label{tab:loss}
    \resizebox{\linewidth}{!}{%
    \begin{threeparttable}%
    \begin{tabular}{c|ccc|c|c}
        \hline
        \multirow{2}*{No.} & \multicolumn{3}{c|}{MIM} & \multirow{2}*{UFM}
            & \multirow{2}*{mIoU} \\ \cline{2-4}
        & Top & Middle & Bottom & & \\ \hline
        1 & S & S &   & E  & 51.8 \\ \hline
        2 &   & S &   & E  & 50.3 \\
        3 & S &   &   & E  & 50.2 \\
        4 & S & S &   &    & 50.5 \\
        5 &   &   &   & E  & 49.0 \\
        6 &   &   &   &    & 48.4 \\ \hline
        7 & S & S & S & E  & 50.9 \\
        8 & S & S &   & S  & 50.8 \\ \hline
        9 & S & S &   & E(FL) & 49.4 \\ \hline
    \end{tabular}
    \begin{tablenotes}
    \item[*] ``S'' denotes semantic supervision.
    ``E'' denotes edge supervision. 
    ``(FL)'' indicates that the focal loss is used for edge supervision.
    \end{tablenotes}
    \end{threeparttable}}
    \vspace{-4mm}
\end{table}

\begin{table}[t]
    \centering
    \renewcommand{\tabcolsep}{3.8mm}
    \caption{Impact of $\lambda$ in \equref{equ:loss}.}
    \label{tab:ablation_lambda}
    \resizebox{.9\linewidth}{!}{%
    \begin{tabular}{c|cccc} \hline
        $\lambda$ & 0 & 0.01 & 0.1 & 1.0 \\ \hline
        mIoU & 49.0 & 49.3 & 51.8 & 51.4 \\ \hline
    \end{tabular}}
    \vspace{-5mm}
\end{table}

\subsection{Booster Training Strategy}\label{sec:super}
Adding supervision to intermediate parts of a network \cite{lee2015deeply} has been shown to be useful for network optimization \cite{zhao2017pyramid,zhang2018context,liu2019richer,liu2018semantic}. In this work, we also use intermediate supervision, however, we propose two different types of supervision. Similar to balancing the computational resources across scales, it is very important to use the right supervision for the right part of the network.

\begin{figure*}[!t]
    \centering
    \subfloat[Input Scan]{%
    \includegraphics[width=0.24\linewidth]{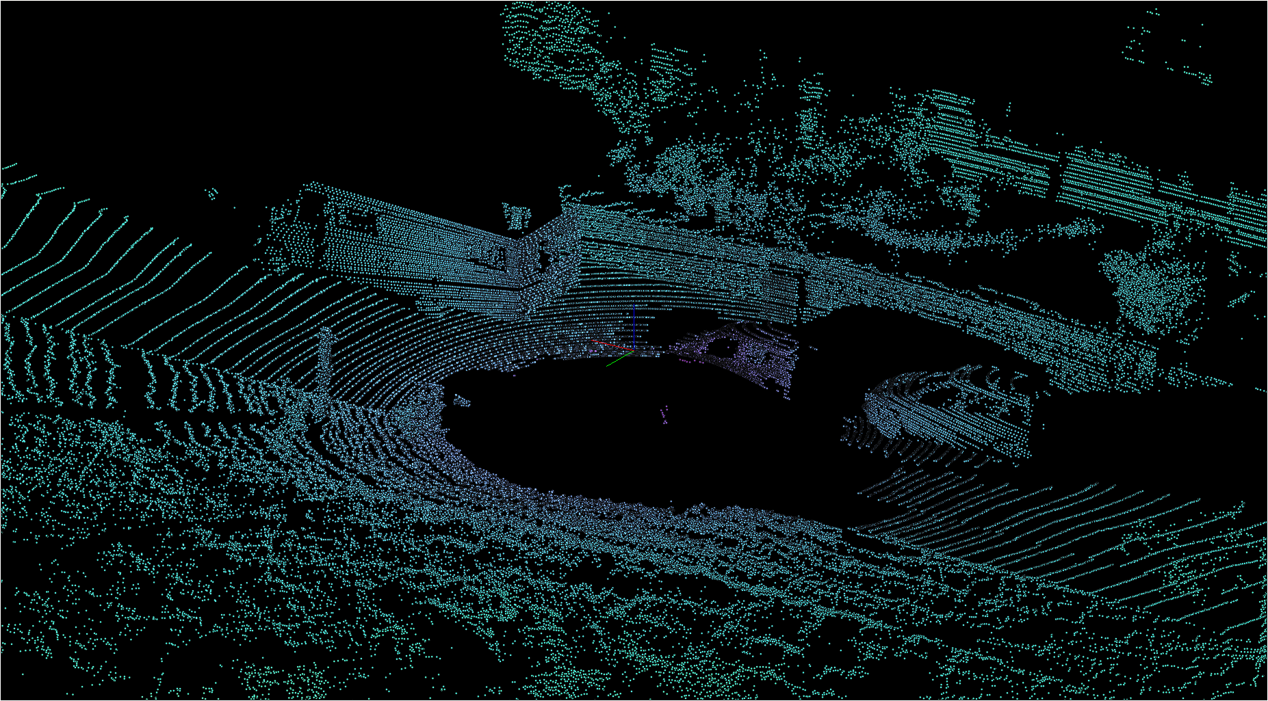}} \hspace{1mm}
    \subfloat[Ground Truth]{%
    \includegraphics[width=0.24\linewidth]{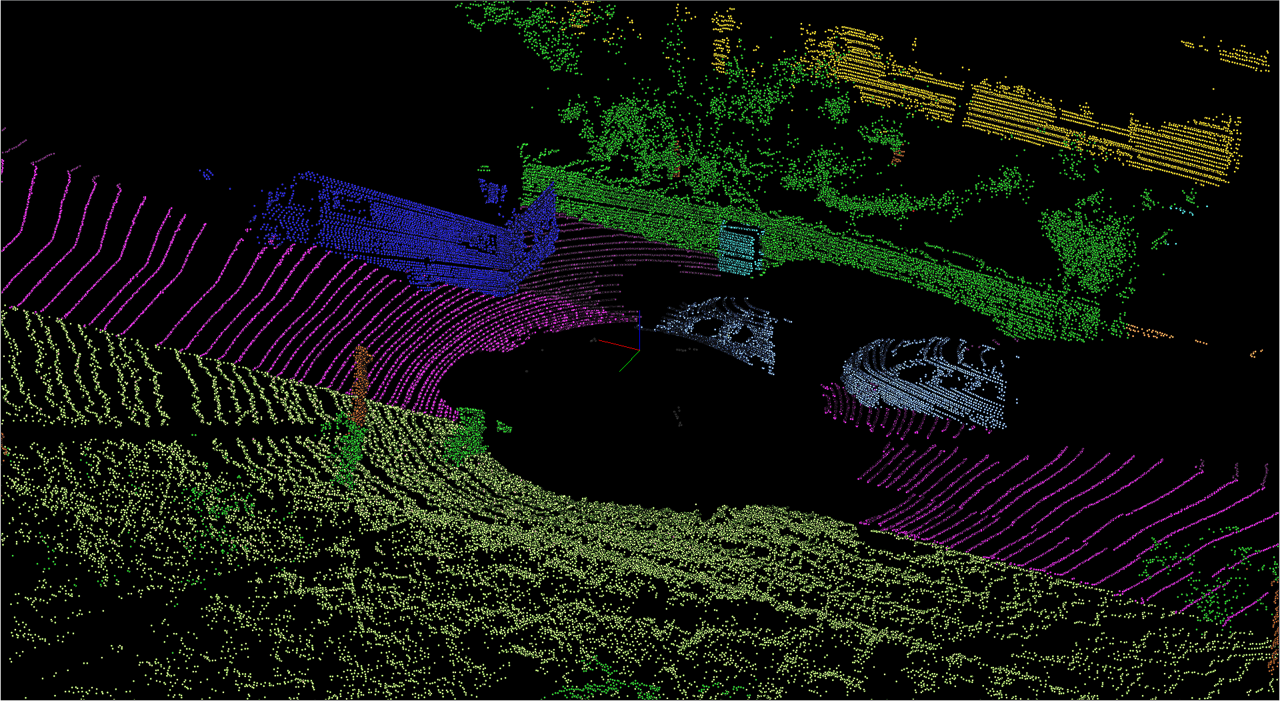}} \hspace{1mm}
    \subfloat[RangeNet53 + $k$-NN]{%
    \includegraphics[width=0.24\linewidth]{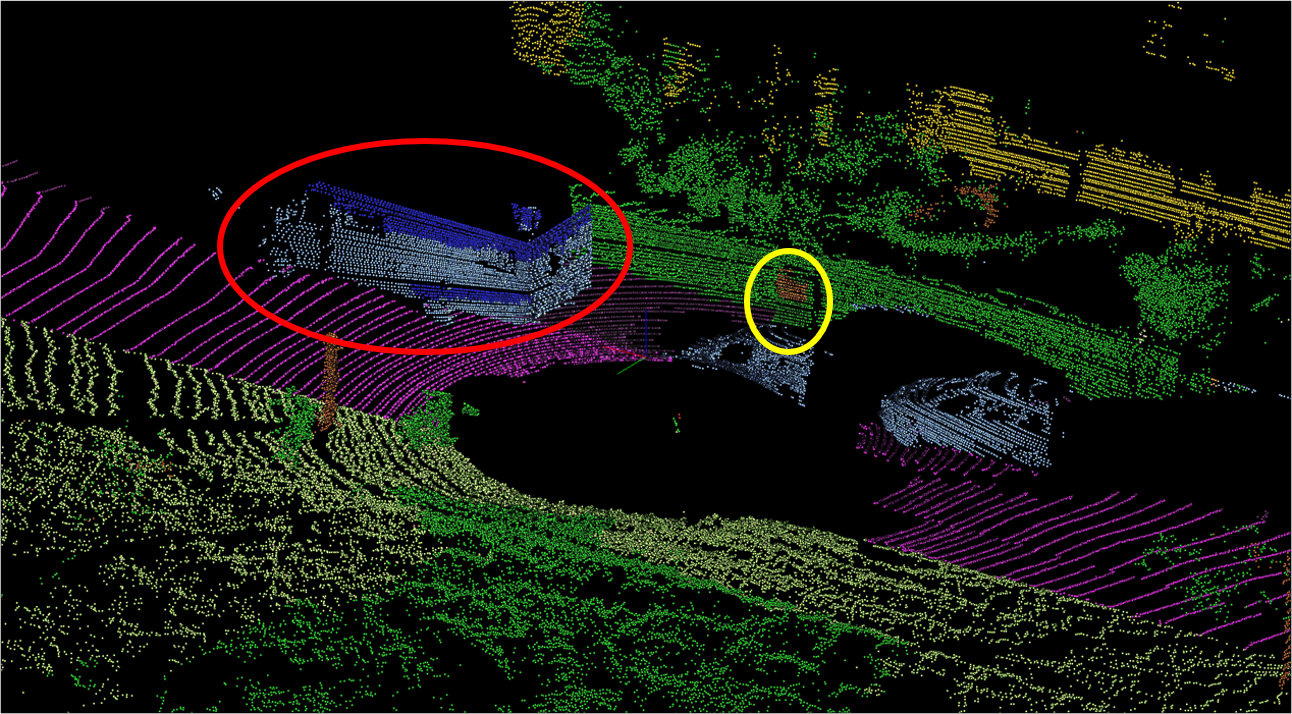}} \hspace{1mm}
    \subfloat[MINet + $k$-NN]{%
    \includegraphics[width=0.24\linewidth]{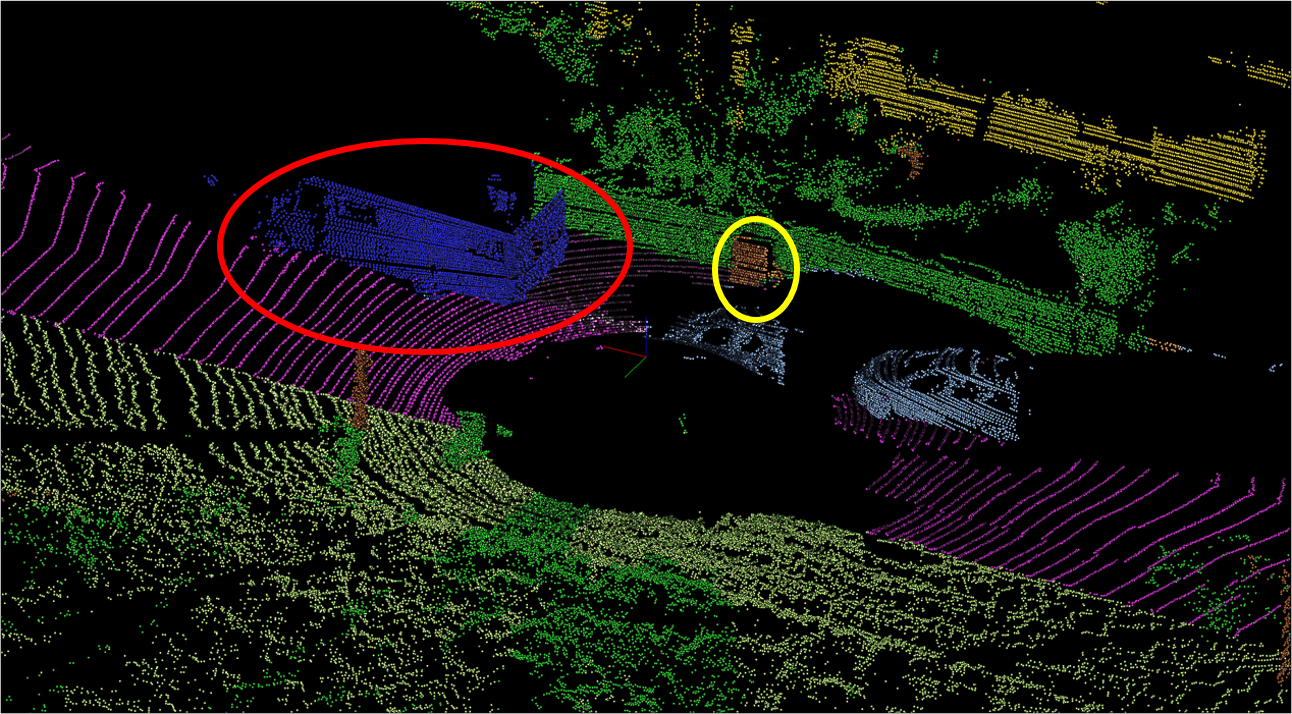}} \hspace{1mm}
    \caption{Qualitative results.
    RangeNet segments most points of the truck as a car, 
    while MINet segments the truck correctly (red circle). 
    In some cases, both approaches fail (yellow circle). 
    Although MINet segments the object correctly, it misclassifies it.}
    \label{fig:vis_3d}
    \vspace{-6mm}
\end{figure*}

We use the standard weighted cross-entropy loss as semantic supervision
\begin{equation}\label{eq:S}
    \mathcal{L}_s = -\frac{1}{|I|}\sum_{i\in I}\sum_{n=1}^{N}w_n p_i^n \log(\hat{p}_i^n),
\end{equation}
where $N$ is the number of classes, $|I|$ is the total number of image pixels, $p_i^n$ is the ground-truth semantic label for pixel $i$ and class $n$ ($p_i^n \in \{0,1\}$), and $\hat{p}_i^n$ is the predicted class probability. The weight $w_n$ for class $n$ is inversely proportional to its occurrence frequency as in \cite{milioto2019rangenet++}.

Besides at the end of the network, we use the weighted cross-entropy loss $\mathcal{L}_{s}$ for the top and middle path as indicated by the dashed arrows in \figref{fig:arch}. As we will show in the experiments, this intermediate supervision improves the training and boosts the accuracy. Adding this semantic supervision to the bottom path, however, does not help since the resolution of the lower path is too low and downsampling of the ground-truth introduces too many artifacts.

As discussed in \secref{sec:ufm}, obtaining accurate segment boundaries after upsampling is an issue. 
Inspired by \cite{yu2018learning,kirillov2017instancecut}, we extract the semantic boundaries from the ground-truth labels and compare it to the semantic boundaries after upsampling. The semantic edge loss is then obtained by
\begin{equation}\label{eq:E}
    \mathcal{L}_e = -\frac{1}{|I|}\sum_{i\in I}\left(e_i \log(\hat{e}_i) + (1-e_i)\log(1-\hat{e}_i)\right),
\end{equation}
where $e_i$ is the ground-truth edge label at pixel $i$ ($e_i\in\{0,1\}$)
and $\hat{e}_i$ is the predicted edge probability at pixel $i$.

Besides of the weighted cross entropy loss, which we denote by $\mathcal{L}_\textit{\text{fs}}$ and which is computed in the same way as $\mathcal{L}_{s}$, we use the Lov\'asz-Softmax loss $\mathcal{L}_\textit{\text{ls}}$ \cite{berman2018lovasz} at the end of the network, which maximizes the intersection-over-union (IoU) score:
\begin{equation}
\mathcal{L}_{l s}=\frac{1}{N} \sum_{n=1}^{N} \overline{\Delta_{J_{n}}}(m(n)),
\end{equation}
\begin{equation}
m_{i}(n)=\left\{\begin{array}{ll}
1 - \hat{p}_i^n & \text{if } p_i^n=1 \\
\hat{p}_i^n & \text{otherwise},
\end{array}\right.
\end{equation}
where $\overline{\Delta_{J_{n}}}$ defines the Lov\'asz extension of the Jaccard index, $\hat{p}_i^n \in [0, 1]$  and $p_i^n \in \{0, 1\}$  denote for class $n$ at pixel $i$ the predicted probability and ground-truth label, respectively.
In summary, the combined loss is given by
\begin{equation}
    \mathcal{L} = \mathcal{L}_\textit{\text{fs}} + \mathcal{L}_\textit{\text{ls}} + \mathcal{L}_{e} + \lambda\sum_s \mathcal{L}_{s}
\label{equ:loss}
\end{equation}
where $\lambda=0.1$ and $\mathcal{L}_{s}$ are the loss functions for the top and middle path. 
\todo{The arrows in \figref{fig:arch} show where each type of supervision is applied.}

\begin{table}[t]
    \centering
    \renewcommand{\tabcolsep}{1.6mm}
    \caption{Ablation study for different blocks of each path.}
    \label{tab:ablation_block}
    \resizebox{\linewidth}{!}{%
    \begin{threeparttable}%
    \begin{tabular}{ccc|cccc} \hline
        Top & Middle & Bottom & Param (M) & GFLOP & SPS & mIoU \\ \hline
        3$\times$MB & 5$\times$MB & 3$\times$BB & 1.0 &  6.20 & 59 & 51.8 \\ \hline
        5$\times$MB & 5$\times$MB & 3$\times$BB & 1.1 &  6.43 & 52 & 50.1 \\
        7$\times$MB & 5$\times$MB & 3$\times$BB & 1.1 &  6.62 & 51 & 49.7 \\
        9$\times$MB & 5$\times$MB & 3$\times$BB & 1.2 &  7.77 & 48 & 48.9 \\ \hline
        3$\times$MB & 3$\times$MB & 3$\times$BB & 1.0 &  6.20 & 60 & 50.8 \\
        3$\times$MB & 7$\times$MB & 3$\times$BB & 1.0 &  6.29 & 55 & 49.0 \\
        3$\times$MB & 9$\times$MB & 3$\times$BB & 1.2 &  6.57 & 51 & 50.2 \\ \hline
        3$\times$MB & 5$\times$MB & 5$\times$BB & 1.4 &  6.43 & 54 & 50.3 \\
        3$\times$MB & 5$\times$MB & 7$\times$BB & 1.8 &  6.62 & 52 & 51.5 \\
        3$\times$MB & 5$\times$MB & 9$\times$BB & 2.4 &  6.92 & 50 & 49.8 \\ \hline
        3$\times$MB & 3$\times$MB & 3$\times$MB & 0.6 & 6.00 & 61 & 49.7 \\
        5$\times$MB & 5$\times$MB & 5$\times$MB & 0.6 &  6.30 & 58 & 50.2 \\
        3$\times$BB & 3$\times$BB & 3$\times$BB & 2.0 & 11.04 & 33 & 51.2 \\ \hline
        \hline
    \end{tabular}
    \begin{tablenotes}
    \item[*] ``MB'' denotes MobileBlock. ``BB'' denotes BasicBlock. 
    ``SPS'' represents the number of processed scans per second.
    \end{tablenotes}
    \end{threeparttable}}
    \vspace{-6mm}
\end{table}

\section{Experiments}

\subsection{Experimental Settings}
We use two challenging datasets to evaluate our method, namely SemanticKITTI \cite{behley2019semantickitti,SemKITTI} and SemanticPOSS\cite{pan2020semanticposs}.
Based on the KITTI Odometry Benchmark~\cite{geiger2012we}, SemanticKITTI provides a semantic label for each point in all scans.
It includes over 43,000 scans from 21 sequences, among which sequences 00 to 10 with over 21,000 scans are available for training and the remaining scans from sequences 11 to 21 are used for testing.
Sequence 08 is used as the validation set, and we train our approach on the remaining training set. We report the results on the validation set for the ablation study. For the test set, which we use to compare to the state-of-the-art, the ground-truth is withheld and the results are evaluated by an on-line server.

\begin{figure}[!t]
    \centering
    \includegraphics[width=\linewidth]{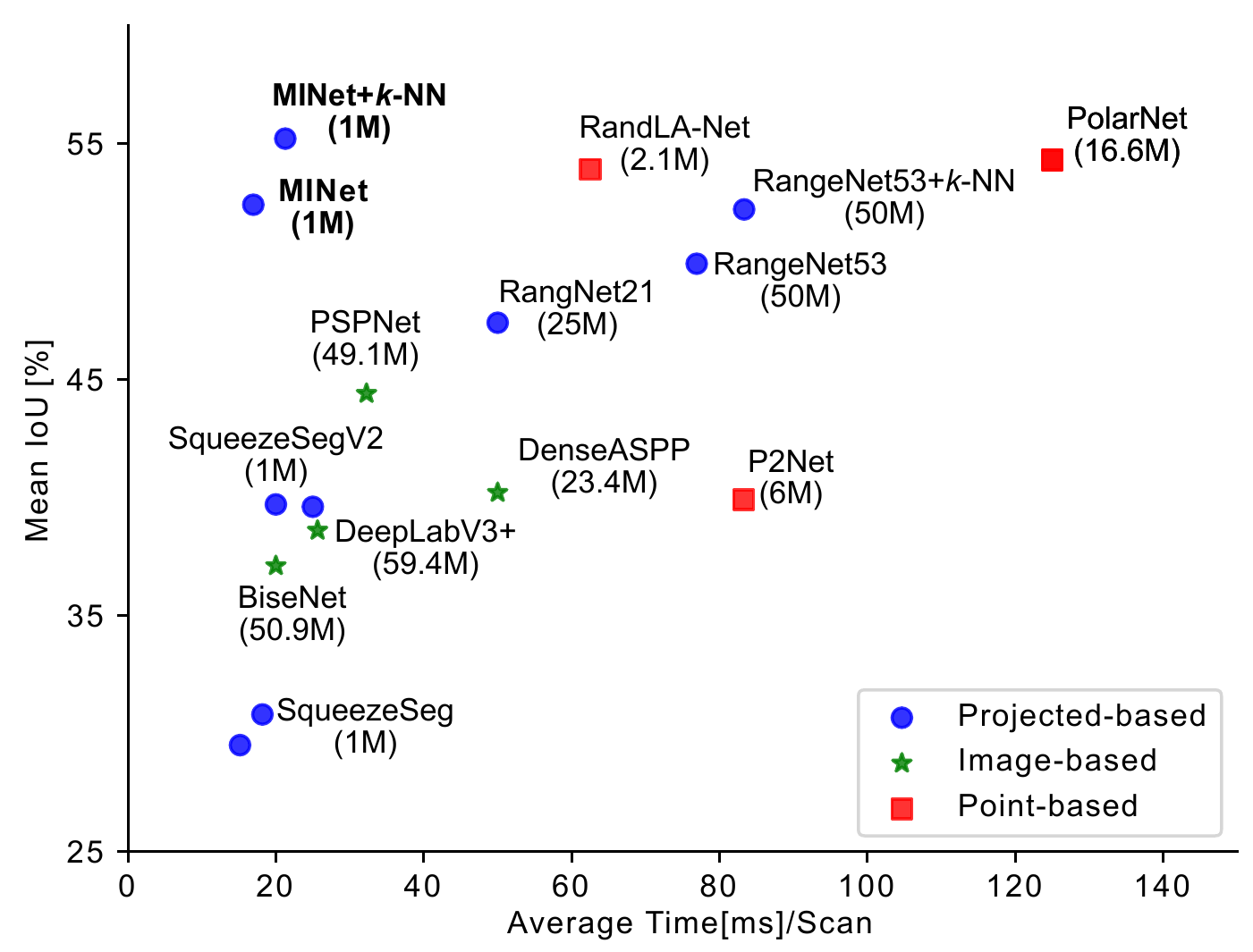}
    \caption{Visualization of \tabref{tab:single_scan}.
    We omit point-based methods that need more than 140ms per scan.
    The proposed MINet is not only the most accurate method,
    it also achieves the best trade-off between accuracy and runtime.}
    \label{fig:runtime}
    \vspace{-6mm}
\end{figure}

\begin{table*}[!t]
    \centering
    \renewcommand{\tabcolsep}{1.2mm}
    \caption{Evaluation results on the SemanticKITTI test set for 
    point-based, image-based, and projection-based methods.}
    \resizebox{0.9\linewidth}{!}{%
    \begin{threeparttable}%
    \begin{tabular}{l|rrrrrrrrrrrrrrrrrrr|c|r|c} \hline
        \diagbox[height=0.63in,width=1.35in]{\raisebox{2.0\height}{Methods}}{\raisebox{-3.5\height}{Classes}}
        & \multicolumn{1}{c}{\rotatebox{90}{\hspace{-0.30in} car}} 
        & \multicolumn{1}{c}{\rotatebox{90}{\hspace{-0.30in} bicycle}}
        & \multicolumn{1}{c}{\rotatebox{90}{\hspace{-0.30in} motorcycle}} 
        & \multicolumn{1}{c}{\rotatebox{90}{\hspace{-0.30in} truck}}
        & \multicolumn{1}{c}{\rotatebox{90}{\hspace{-0.30in} other-vehicle}}
        & \multicolumn{1}{c}{\rotatebox{90}{\hspace{-0.30in} person}}
        & \multicolumn{1}{c}{\rotatebox{90}{\hspace{-0.30in} bicyclist}}
        & \multicolumn{1}{c}{\rotatebox{90}{\hspace{-0.30in} motorcyclist}}
        & \multicolumn{1}{c}{\rotatebox{90}{\hspace{-0.30in} road}}
        & \multicolumn{1}{c}{\rotatebox{90}{\hspace{-0.30in} parking}}
        & \multicolumn{1}{c}{\rotatebox{90}{\hspace{-0.30in} sidewalk}}
        & \multicolumn{1}{c}{\rotatebox{90}{\hspace{-0.30in} other-ground}}
        & \multicolumn{1}{c}{\rotatebox{90}{\hspace{-0.30in} building}}
        & \multicolumn{1}{c}{\rotatebox{90}{\hspace{-0.30in} fence}}
        & \multicolumn{1}{c}{\rotatebox{90}{\hspace{-0.30in} vegetation}}
        & \multicolumn{1}{c}{\rotatebox{90}{\hspace{-0.30in} trunk}}
        & \multicolumn{1}{c}{\rotatebox{90}{\hspace{-0.30in} terrain}}
        & \multicolumn{1}{c}{\rotatebox{90}{\hspace{-0.30in} pole}}
        & \multicolumn{1}{c|}{\rotatebox{90}{\hspace{-0.30in} traffic-sign}} 
        & \multicolumn{1}{c|}{\rotatebox{90}{\hspace{-0.30in} SPS}}
        & \multicolumn{1}{c|}{\rotatebox{90}{\hspace{-0.30in} Param (M)}} 
        & \multicolumn{1}{c}{\rotatebox{90}{\hspace{-0.30in} mIoU}} \\ \hline
        Pointnet \cite{qi2017pointnet} & 46.3 & 1.3 & 0.3 & 0.1 & 0.8 & 0.2 & 0.2 & 0.0 & 61.6 & 15.8 & 35.7 & 1.4 & 41.4 & 12.9 & 31.0 & 4.6 & 17.6 & 2.4 & 3.7 & 2 & 3.0 & 14.6\\ 
        Pointnet++ \cite{qi2017pointnet++} & 53.7 & 1.9 & 0.2 & 0.9 & 0.2 & 0.9 & 1.0 & 0.0 & 72.0 & 18.7 & 41.8 & 5.6 & 62.3 & 16.9 & 46.5 & 13.8 & 30.0 & 6.0 & 8.9 & 0.1 & 6.0 & 20.1\\
        SPGraph \cite{landrieu2018large} & 68.3 & 0.9 & 4.5 & 0.9 & 0.8 & 1.0 & 6.0 & 0.0 & 49.5 & 1.7 & 24.2 & 0.3 & 68.2 & 22.5 & 59.2 & 27.2 & 17.0 & 18.3 & 10.5 & 0.2 & \textbf{0.3} & 20.0 \\
        SPLATNet \cite{su2018splatnet} & 66.6 & 0.0 & 0.0 & 0.0 & 0.0 & 0.0 & 0.0 & 0.0 & 70.4 & 0.8 & 41.5 & 0.0 & 68.7 & 27.8 & 72.3 & 35.9 & 35.8 & 13.8 & 0.0 & 1 & 0.8 & 22.8 \\
        TangentConv \cite{tatarchenko2018tangent} & 86.8 & 1.3 & 12.7 & 11.6 & 10.2 & 17.1 & 20.2 & 0.5 & 82.9 & 15.2 & 61.7 & 9.0 & 82.8 & 44.2 & 75.5 & 42.5 & 55.5 & 30.2 & 22.2 & 0.3 & 0.4 & 35.9\\
        P$^2$Net \cite{li2020projected} & 85.6 & 20.4 & 14.4 & 14.4 & 11.5 & 16.9 & 24.9 & 5.9 & 87.8 & 47.5 & 67.3 & 7.3 & 77.9 & 43.4 & 72.5 & 36.5 & 60.8 & 22.8 & 38.2 & 12 & 6.0 & 39.8 \\
        RandLA-Net \cite{hu2019randla} & \textbf{94.2} & 26.0 & 25.8 & \textbf{40.1} & \textbf{38.9} & 49.2 &  48.2 & 7.2 & 90.7 & 60.3 & 73.7 & 20.4 & 86.9 & 56.3 & 81.4 & 61.3 & 66.8 & 49.2 & 47.7 & 16 & 2.1 & 53.9\\
        PolarNet \cite{zhang2020polarnet} & 93.8 & 40.3 & 30.1 & 22.9 & 28.5 & 43.2 & 40.2 & 5.6 & 90.8 & 61.7 & 74.4 & 21.7 & \textbf{90.0} & \textbf{61.3} & \textbf{84.0} & \textbf{65.5} & \textbf{67.8} & \textbf{51.8} & 57.5 & 8 & 16.6  & 54.3 \\ \hline
        DeepLabV3+\cite{chen2018encoder} & 78.4 & 13.6 & 9.5 & 9.5 & 10.4 & 17.5 & 22.0 & 0.4 & 88.5 & 54.5 & 66.7 & 9.7 & 77.9 & 39.1 & 72.0 & 39.9 & 60.0 & 23.4 & 36.1 & 39 & 59.4 & 38.4 \\
        PSPNet\cite{li2018pyramid} & 79.6 & 25.0 & 26.4 & 17.5 & 24.0 & 34.1 & 28.4 & 7.3 & 90.2 & 58.2 & 70.2 & 19.9 & 79.7 & 43.5 & 74.2 & 43.2 & 61.2 & 23.1 & 37.5 & 31 & 49.1 & 44.4\\
        BiSeNet\cite{yu2018bisenet} &76.0 & 13.4 & 11.9 & 18.3 & 7.6 & 16.4 & 26.0 & 0.5 & 87.6 & 49.9 & 64.2 & 6.5 & 74.7 & 34.7 & 69.7 & 36.8 & 58.0 & 19.6 & 32.3 & 50 & 50.9 & 37.1 \\
        DenseASPP\cite{yang2018denseaspp} & 78.1 & 20.5 & 18.2 & 20.0 & 16.6 & 27.8 & 28.9 & 5.7 & 88.5 & 53.3 & 67.5 & 9.3 & 76.3 & 39.6 & 70.0 & 36.8 & 57.7 & 15.9 & 32.4 & 20 & 23.4 & 40.2\\ \hline
        SqueezeSeg \cite{wu2018squeezeseg} & 68.8 & 16.0 & 4.1 & 3.3 & 3.6 & 12.9 & 13.1 & 0.9 & 85.4 & 26.9 & 54.3 & 4.5 & 57.4 & 29.0 & 60.0 & 24.3 & 53.7 & 17.5 & 24.5 & \textbf{66} & 1.0 & 29.5\\
        SqueezeSeg + CRF \cite{wu2018squeezeseg} & 68.3 & 18.1 & 5.1 & 4.1 & 4.8 & 16.5 & 17.3 & 1.2 & 84.9 & 28.4 & 54.7 & 4.6 & 61.5 & 29.2 & 59.6 & 25.5 & 54.7 & 11.2 & 36.3 & 55 & 1.0 & 30.8 \\
        SqueezeSegV2 \cite{wu2019squeezesegv2} & 81.8 & 18.5 & 17.9 & 13.4 & 14.0 & 20.1 & 25.1 & 3.9 & 88.6 & 45.8 & 67.6 & 17.7 & 73.7 & 41.1 & 71.8 & 35.8 & 60.2 & 20.2 & 36.3 & 50 & 1.0 &  39.7\\
        SqueezeSegV2 + CRF \cite{wu2019squeezesegv2} & 82.7 & 21.0 & 22.6 & 14.5 & 15.9 & 20.2 & 24.3 & 2.9 & 88.5 & 42.4 & 65.5 & 18.7 & 73.8 & 41.0 & 68.5 & 36.9 & 58.9 & 12.9 & 41.0 & 40 & 1.0 & 39.6 \\
        RangeNet21 \cite{milioto2019rangenet++} & 85.4 & 26.2 & 26.5 & 18.6 & 15.6 & 31.8 & 33.6 & 4.0 & 91.4 & 57.0 & 74.0 & 26.4 & 81.9 & 52.3 & 77.6 & 48.4 & 63.6 & 36.0 & 50.0 & 20 & 25.0 & 47.4 \\
        RangeNet53 \cite{milioto2019rangenet++} & 86.4 & 24.5 & 32.7 & 25.5 & 22.6 & 36.2 & 33.6 & 4.7 & \textbf{91.8} & 64.8 & 74.6 & \textbf{27.9} & 84.1 & 55.0 & 78.3 & 50.1 & 64.0 & 38.9 & 52.2 & 13 & 50.4 & 49.9\\
        RangeNet53 + $k$-NN \cite{milioto2019rangenet++} & 91.4 & 25.7 & \textbf{34.4} & 25.7 & 23.0 & 38.3 & 38.8 & 4.8 & \textbf{91.8} & \textbf{65.0} & \textbf{75.2} & 27.8 & 87.4 & 58.6 & 80.5 & 55.1 & 64.6 & 47.9 & 55.9 & 12 & 50.4 & 52.2 \\ \hline
        MINet & 85.2 & 38.2 & 32.1 & 29.3 & 23.1 & 47.6 & 46.8 & 24.5 & 90.5 & 58.8 & 72.1 & 25.9 & 82.2 & 49.5 & 78.8 & 52.5 & 65.4 & 37.7 & 55.5 & 59 & 1.0 & 52.4\\
        MINet + $k$-NN & 90.1 & \textbf{41.8} & 34.0 & 29.9 & 23.6 & \textbf{51.4} & \textbf{52.4} & \textbf{25.0} & 90.5 & 59.0 & 72.6 & 25.8 & 85.6 & 52.3 & 81.1 & 58.1 & 66.1 & 49.0 & \textbf{59.9} & 47 & 1.0 &  \textbf{55.2} \\ \hline
    \end{tabular}
    \begin{tablenotes}
    \item[*] The proposed MINet performs well for small objects.
    \end{tablenotes}
    \end{threeparttable}}
    \label{tab:single_scan}
\vspace{-6mm}
\end{table*}

SemanticPOSS is a smaller dataset with 2988 LiDAR scans, captured at the Peking University. 
The point clouds of SemanticPOSS are more sparse compared to SemanticKITTI due to the lower resolution of the LiDAR sensor. 
SemanticPOSS is split into six subsets equally, among which we use the 3$^{\rm rd}$ subset for validation and the others for training.
We report the results on the validation set.
Since these two datasets differ in size, LiDAR sensor, and environment, they provide an ideal testbed for evaluating the proposed approach.    
As for the evaluation metric, we calculate the standard mean intersection over union (mIoU)
\cite{everingham2015pascal} over all classes:
\begin{equation}
    {\rm mIoU} = \frac{1}{N} \sum_{n=1}^{N} \frac{{\rm TP}_n}{{\rm TP}_n + {\rm FP}_n + {\rm FN}_n}
\end{equation}
where ${\rm TP}_n$, ${\rm FP}_n$, and ${\rm FN}_n$ denote the numbers of true
positive, false positive, and false negative predictions for
class $n$, respectively.
$N$ is the number of classes.


\subsection{Ablation Study}

\subsubsection{Modules}\label{sec:modules}
As illustrated in \figref{fig:arch}, our network consists of three modules.
In \tabref{tab:ablation} we evaluate some design choices for the Mini Fusion Module (MFM) (\secref{sec:mfm}), the Multi-scale Interaction Module (MIM) (\secref{sec:mim}), and the Up Fusion Module (UFM) (\secref{sec:ufm}). 
While the proposed approach achieves 51.8\% mIoU (Row 4), we evaluate the impact of processing each modality separately before fusing them in the features space in the first row. To keep the number of parameters the same, we use 3$\times$3 convolutions to process the input multi-modal image instead of processing each modality separately in MFM. In this case, the accuracy drops by 0.9\% (Row 1). This shows the benefit of processing each modality separately at the beginning. \todo{In the last row, we also report the result when we omit the depth from the input. In this case, the accuracy drops to 2.2\%.} In \figref{fig:arch}, we have connections between the three paths. If we remove the top-to-down interactions, the accuracy is reduced by 1.1\% (Row 2). This demonstrates the benefit of allowing interactions between the multi-scale features. If the multi-resolution features are just resized, concatenated, and processed by convolutional layers, instead of using UFM, the accuracy is reduced by 1.2\% (Row 3).

\subsubsection{Supervision Setting}
As discussed in \secref{sec:super}, we use additional supervision for MIM and UFM. For MIM, we use the semantic labels (S) to add the loss\equref{eq:S} to the top and middle path. For UFM, we add the loss\equref{eq:E} for the segment boundaries (E). In \tabref{tab:loss}, we report the mIoU for different settings. The best setting is achieved by adding semantic supervision to the top and middle path in MIM and applying edge supervision to UFM (Row 1). Removing any of this additional supervision leads to an accuracy loss by more than 1.3\% (Row 2-4). If the additional supervision is only used for UFM, the accuracy even decreases further (Row 5). If we do not use any additional supervision, the accuracy is lowest and 3.4\% below the proposed setting (Row 6). While we removed so far additional supervision, the last two rows in the table show results when we add or change the type of supervision. If we add additional supervision to the bottom path, the accuracy decreases by 0.9\%. This is due to the large difference between the ground-truth resolution and the resolution of the bottom path, which results in sampling artifacts that have a negative impact.   
Instead of using the edge loss\equref{eq:E} for UFM, we also replaced it by the semantic loss that is used for MIM. While adding the semantic loss (Row 8) is better than using no additional loss for UFM (Row 4), the edge loss (Row 1) achieves a 1\% higher accuracy than the semantic loss.
This is expected
since the purpose of the edge loss is to improve the segment
boundaries after the upscaling, which is done by the Up
Fusion Module.
We also investigated what happens if the focal loss \cite{lin2017focal} is used for the edge loss instead of\equref{eq:E} (Row 9). In this case, the accuracy decreases.

\subsubsection{Impact of $\lambda$}
Our loss function\equref{equ:loss} contains only one hyper-parameter, namely $\lambda$. We evaluate the impact of $\lambda$ in \tabref{tab:ablation_lambda}. The setting $\lambda=0$ corresponds to row 5 of \tabref{tab:loss} where no additional supervision is added to the paths. Setting $\lambda$ between $0.1$ and $1.0$ performs well.

\begin{table*}[!t]
    \centering
    \caption{Evaluation results on the SemanticPOSS dataset. }
    \resizebox{0.8\linewidth}{!}{%
    \begin{tabular}{c|ccccccccccc|c}
    \hline
           & 
         {person} & 
         {rider} & 
         {car} & 
         {trunk} & 
         {plants} & 
         {traffic sign} & 
         {pole} & 
         {building} & 
         {fence} & 
         {bike} &
         {road} &
         {mIoU} \\
         \hline
         SqueezeSegV1\cite{wu2018squeezeseg} & 5.5&0.0&8.7&3.4&39.1&2.4&2.5&34.5&7.6&18.4&62.5&16.8 \\ 
         SqueezeSegV1\cite{wu2018squeezeseg} + CRF & 14.2&1.4&11.6&18.1&5.9&11.1&1.9&37.9&5.6&18.9&78.7&18.7\\
         SqueezeSegV2\cite{wu2019squeezesegv2} & 18.4 & 11.2 & 34.9 & 15.8 & 56.3 & 11.0 & 4.5 & 47.0 & 25.5 & 32.4 & 71.3  & 29.8\\
         SqueezeSegV2\cite{wu2019squeezesegv2} + CRF & \textbf{23.9}&\textbf{22.6}&29.7&15.3&37.3&11.1&\textbf{5.3}&45.9&18.2&34.7&\textbf{73.4}&28.9\\
         RangeNet53\cite{milioto2019rangenet++} & 10.0&6.2&33.4&7.3&54.2&5.5&2.6&49.9&18.4&28.6&63.5&25.4 \\ 
         RangeNet53 + $k$-NN & 14.2&8.2&35.4&9.2&58.1&6.8&2.8&55.5&\textbf{28.8}&32.2&66.3&28.9\\
        \hline
         MINet & 13.3 &11.3 &34.0 & 18.8 & 62.9 &11.8 &4.1 &55.5 &20.4 &34.7 &69.2 & 30.5\\
         MINet + $k$-NN & {20.1}&15.1&\textbf{36.0}&\textbf{23.4}&\textbf{67.4}&\textbf{15.5}&{5.1}&\textbf{61.6}&{28.2}&\textbf{40.2}&72.9&\textbf{35.1}\\
         \hline
    \end{tabular}}
    \vspace{-6mm}
    \label{tab:single_scan_poss}
\end{table*}

\begin{table}[!t]
    \centering
    \renewcommand{\tabcolsep}{2.0mm}
    \caption{Performance on an embedded platform (Jetson AGX).}
    \label{tab:mobile}
    \resizebox{0.9\linewidth}{!}{%
    \begin{threeparttable}%
    \begin{tabular}{l|l|rrr} \hline
        Methods & Resolutions & GFLOPs & SPS & mIoU \\ \hline
        \multirow{3}*{RangeNet53 \cite{milioto2019rangenet++}} 
            & 64 $\times$ 2048 & 360.5 &  7 & 49.9 \\
            & 64 $\times$ 1024 & 180.3 & 11 & 45.4 \\
            & 64 $\times$ 512  &  90.1 & 22 & 39.3 \\ \hline
        \multirow{3}*{RangeNet53 + $k$-NN \cite{milioto2019rangenet++}}
            & 64 $\times$ 2048 & 360.5 &  5 & 52.2 \\
            & 64 $\times$ 1024 & 180.3 &  8 & 48.0 \\
            & 64 $\times$ 512  &  90.1 & 13 & 41.9 \\ \hline\hline
        \multirow{3}*{MINet}
            & 64 $\times$ 2048 & 6.2 & 24 & 52.4 \\
            & 64 $\times$ 1024 & 3.2 & 47 & 49.1 \\
            & 64 $\times$  512 & 1.7 & 80 & 45.0 \\ \hline
        \multirow{3}*{MINet + $k$-NN}
            & 64 $\times$ 2048 & 6.2 & 13 & 55.2 \\
            & 64 $\times$ 1024 & 3.2 & 18 & 52.4 \\
            & 64 $\times$  512 & 1.7 & 21 & 48.5 \\
         \hline
    \end{tabular}
    \begin{tablenotes}
    \item[*] The number of GFLOPs does not include the post-processing.
    \end{tablenotes}
    \end{threeparttable}}
\vspace{-6mm}
\end{table}

\subsubsection{Path Settings}
As shown in \tabref{tab:instantiation}, we balance the computational resources across the three paths where we increase the complexity as the resolution decreases. 
In rows \mbox{1-10} of \tabref{tab:ablation_block}, we report the results when we vary the number of blocks for the top, middle, and bottom path. The results show that increasing the parameters only for one path does not result in an improvement. For instance, using 9 instead of 3 MobileBlocks for the top path (Row 4) decreases the accuracy by $2.9\%$. This shows that a good computational balance between the paths is required. In rows 11-13, we report the results when we use the same operations for all paths, \ie either 3 or 5 MobileBlocks or 3 BasicBlocks.
Using only MobileBlocks reduces the number of parameters, but it improves the runtime only slightly and this is only the case for 3 MobileBlocks. This, however, comes at a substantially lower accuracy. 
In terms of runtime and accuracy, the proposed setting provides a much better trade-off. 
If the computational expensive BasicBlocks are used for all paths, the number of parameters and runtime nearly doubles while the accuracy is nearly the same. 
This shows that using the same operations for all resolutions is highly inefficient and that the proposed approach achieves a good balance between efficiency and effectiveness.

\subsection{Comparison with other Methods}
We first compare the proposed approach (MINet) with other methods on the SemanticKITTI test set in terms of both accuracy and efficiency.
\todo{For a fair comparison, all methods including image-based methods are trained from scratch.}
The results are shown in \tabref{tab:single_scan}. In the first rows, we show the results for point-based methods. Most of the approaches are very slow and cannot process more than 2 scans per second since spatial aggregation operations are usually very time-consuming for large point clouds. 
The very recent works \cite{li2020projected,hu2019randla,zhang2020polarnet} are faster and process up to 16 scans per second, but it is difficult to deploy these networks on embedded systems due to their complex operations. 
The highest accuracy is achieved by \cite{zhang2020polarnet}, but the network is very large with over 16M parameters. Our proposed approach outperforms all point-based methods in terms of runtime, number of parameters, and accuracy. 


%

As for image-based methods, we use four widely used methods, namely PSPNet \cite{zhao2017pyramid}, DeepLabV3+ \cite{chen2018encoder}, DenseASPP \cite{yang2018denseaspp}, and the lightweight model BiseNet \cite{yu2018bisenet}.
We adjust the input channels so that these methods can be applied to the projection map.
While these methods are faster than point-based methods, they have by far more parameters and the accuracy is significantly lower. This shows that projected LiDAR data cannot be directly processed by image-based segmentation methods since the modality differs from RGB images. 

We also compare our approach to other projection-based methods. While SqueezeSeg \cite{wu2018squeezeseg, wu2019squeezesegv2} has the same amount of parameters and depending on the setting runs slightly faster, the accuracy is very low. The \sArt projection-based method RangeNet \cite{milioto2019rangenet++} outperforms SqueezeSeg in terms of accuracy, but this is achieved by increasing the number of parameters to over 50M and decreasing the runtime. Our approach is much more efficient. It uses only 2\% of the number of parameters compared to RangeNet53 and it is about 4$\times$ faster while achieving a higher accuracy. \figref{fig:runtime} visualizes the accuracy and runtime of the methods of \tabref{tab:single_scan} and shows the effectiveness and efficiency of the proposed approach.  

We also evaluate our method on SemanticPOSS \cite{pan2020semanticposs} and compare our approach to other methods in \tabref{tab:single_scan_poss}. Since the dataset is smaller and the point clouds are more sparse compared to SemanticKitti, the mIoU is lower for all methods. However, our approach still outperforms other methods with a large margin.
This proves the effectiveness of our method.

\subsection{Performance on an Embedded Platform}
We finally compare our method with RangeNet on an embedded platform. 
Here, we use a Jetson AGX that is an AI module for embedded systems, as it is usually used for autonomous driving.
\todo{We optimize our method and RangeNet using TensorRT.}
The results are summarized in \tabref{tab:mobile}.
Since the input resolution can be decreased to reduce the runtime, we report the results for three different input resolutions.
We can see that at each resolution, the proposed MINet outperforms RangeNet53 \cite{milioto2019rangenet++} with or without post-processing. Furthermore, MINet is also much faster than RangeNet53.
Specifically, MINet is about 4$\times$ faster than RangeNet53 without post-processing and 2$\times$ faster with post-processing.
Such high efficiency makes MINet quite suitable for robotics applications.
Even with full resolution and post-processing, MINet runs at real-time since the LiDAR scan frequency is 10Hz. 
Compared to \tabref{tab:single_scan} where the runtime is measured on a workstation with a single Quadro P6000, the post-processing has a higher impact on the runtime for the embedded platform since the post-processing is not optimized by TensorRT.
Moreover, the post-processing is applied to the point cloud, so it cannot benefit from reducing the resolution of the projection map.

\section{Conclusion}
In this work, we proposed a novel lightweight projection-based method, called Multi-scale Interaction Network (MINet), for semantic segmentation of LiDAR data. 
The network is highly efficient and runs in real-time on GPUs and embedded platforms. 
It outperforms point-based, image-based, and projection-based methods in terms of accuracy, number of parameters, and runtime. 
This is achieved by using a multi-scale approach where the computational resources are balanced between the scales and by introducing interactions between the scales. 
By processing the modalities separately before fusing them and adding additional different types of supervision, we could further improve the accuracy without decreasing the runtime. 
Compared to the \sArt projection-based method RangeNet, MINet reduces the number of parameters by 98\% and is 4$\times$ faster while achieving a higher accuracy. 
Since MINet processes more than 24 scans per second on an embedded platform, it can be used for autonomous vehicles and robots.
Our method also achieves good performance on other tasks, like moving object segmentation \cite{chen2021ral}.
Projection-based methods, however, have some limitations. For instance, it is not straightforward to integrate temporal information from multiple views. Finally, we expect that the design principles of the network are also valuable for other tasks like 3D car detection. The source code is available at \url{https://github.com/sj-li/MINet}.




\bibliographystyle{IEEEtran}
\bibliography{iros}

\end{document}